\documentclass[letterpaper]{article} 
\usepackage{aaai2026}
\usepackage{times}  
\usepackage{helvet}  
\usepackage{courier}  
\usepackage[hyphens]{url}  
\usepackage{graphicx} 
\usepackage{natbib}  
\usepackage{caption} 
\usepackage{algorithm}
\usepackage{algorithmic}
\usepackage{amsfonts}
\usepackage[table,xcdraw,dvipsnames]{xcolor}
\usepackage{newfloat}
\usepackage{listings}
\usepackage{amsmath}
\usepackage{multirow}
\usepackage{colortbl}
\usepackage{pifont}
\usepackage{subcaption}
\usepackage{booktabs}
\usepackage{nicefrac}
\usepackage{microtype}
\usepackage{amsfonts}
\usepackage{url} 

\DeclareCaptionStyle{ruled}{labelfont=normalfont,labelsep=colon,strut=off}
\floatstyle{ruled}
\newfloat{listing}{tb}{lst}{}
\floatname{listing}{Listing}

\title{Structures Meet Semantics: Multimodal Fusion via Graph Contrastive Learning}
\author{
    Jiangfeng Sun\textsuperscript{\rm 1},
    Sihao He\textsuperscript{\rm 1},
    Zhonghong Ou\textsuperscript{\rm 1},
    Meina Song\textsuperscript{\rm 1}
}
\affiliations{
    \textsuperscript{\rm 1}Beijing University of Posts and Telecommunications, Beijing, China\\
    sun2017@bupt.edu.cn, sihaohe@bupt.edu.cn, zhonghong.ou@bupt.edu.cn, mnsong@bupt.edu.cn
}

\usepackage{bibentry}

\begin{document}

\maketitle

\begin{abstract}
Multimodal sentiment analysis (MSA) aims to infer emotional states by effectively integrating textual, acoustic, and visual modalities. Despite notable progress, existing multimodal fusion methods often neglect modality-specific structural dependencies and semantic misalignment, limiting their quality, interpretability, and robustness. To address these challenges, we propose a novel framework called the Structural-Semantic Unifier (SSU), which systematically integrates modality-specific structural information and cross-modal semantic grounding for enhanced multimodal representations. Specifically, SSU dynamically constructs modality-specific graphs by leveraging linguistic syntax for text and a lightweight, text-guided attention mechanism for acoustic and visual modalities, thus capturing detailed intra-modal relationships and semantic interactions. We further introduce a semantic anchor, derived from global textual semantics, that serves as a cross-modal alignment hub, effectively harmonizing heterogeneous semantic spaces across modalities. Additionally, we develop a multiview contrastive learning objective that promotes discriminability, semantic consistency, and structural coherence across intra- and inter-modal views. Extensive evaluations on two widely-used benchmark datasets, CMU-MOSI and CMU-MOSEI, demonstrate that SSU consistently achieves state-of-the-art performance while significantly reducing computational overhead compared to prior methods. Comprehensive qualitative analyses further validate SSU’s interpretability and its ability to capture nuanced emotional patterns through semantically-grounded interactions.
\end{abstract}

\section{Introduction}
Multimodal sentiment analysis (MSA) aims to automatically infer human emotional states by jointly analyzing complementary signals from textual, acoustic, and visual modalities. With the rapid proliferation of multimedia content and the growing need for emotionally intelligent systems in areas such as affective computing, human-computer interaction, and social media analytics, effective multimodal fusion strategies have become increasingly critical.

Attention-based methods have recently emerged as predominant solutions for multimodal fusion, primarily due to their flexibility in modeling dynamic interactions across modalities. However, most existing fusion techniques treat each modality as a simple sequence of features, applying feature-level interactions through standard query-key-value mechanisms. Such approaches frequently disregard inherent modality-specific structural dependencies, including linguistic syntax in textual data and temporal coherence in audio and visual streams. Additionally, current methods implicitly assume that semantic content across modalities is naturally aligned—a simplification that often breaks down under conditions of ambiguous or nuanced sentiment expression.
\begin{figure}[t!]
    \centering
    \includegraphics[width=1\linewidth]{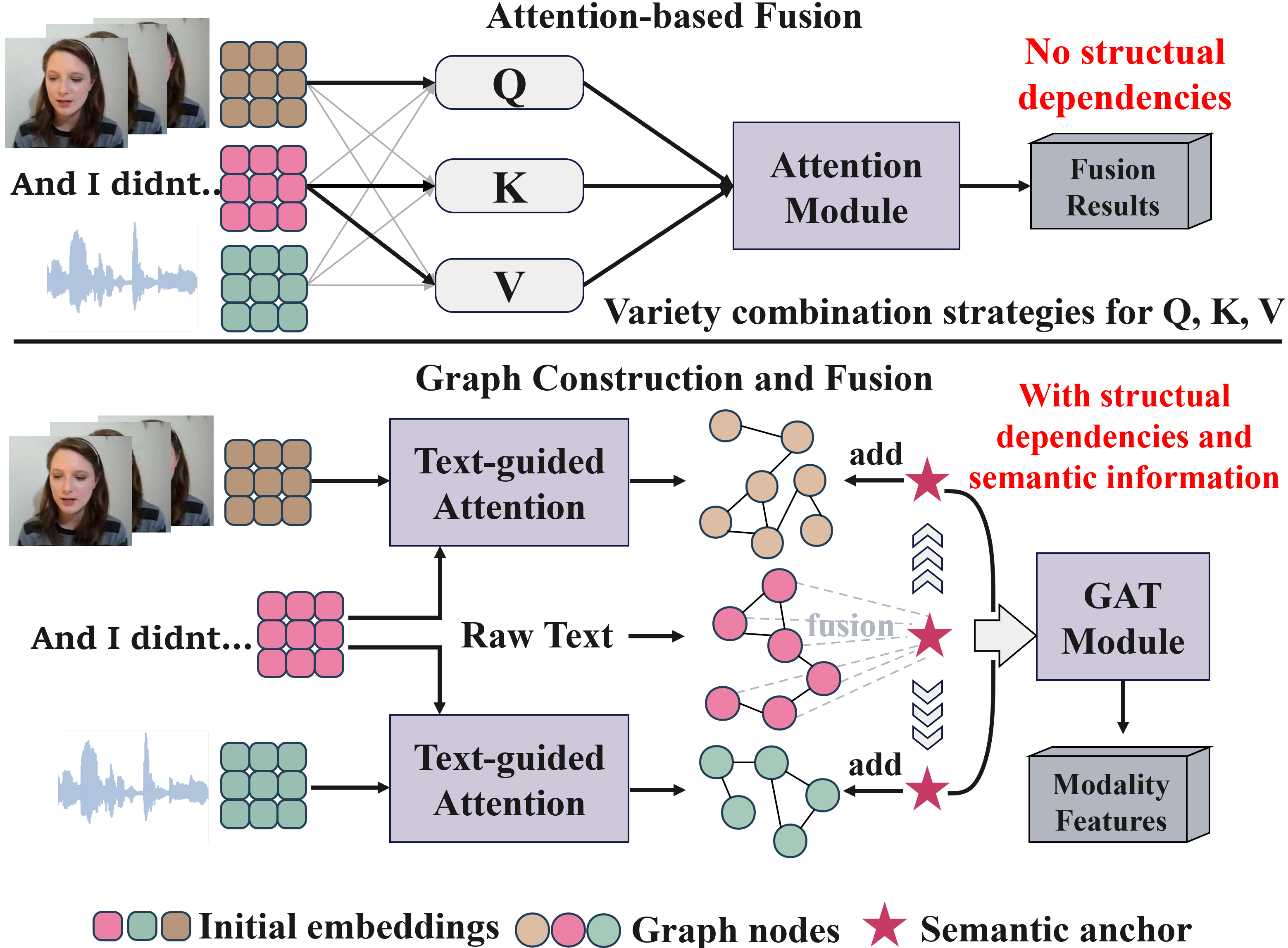}
    \caption{Comparison between attention-based fusion and our proposed SSU-based graph fusion framework.}
    \label{fig:advertisement}
\end{figure}

These fundamental limitations hinder both the representational robustness and interpretability of multimodal sentiment models. Without explicit modeling of structural dependencies or targeted mechanisms to address semantic misalignment across modalities, learned representations tend to be ambiguous, less grounded, and less interpretable. Figure~\ref{fig:advertisement} illustrates these limitations by comparing traditional attention-based fusion with our proposed approach. As highlighted, conventional methods inadequately capture critical intra-modal structural cues and cross-modal semantic coherence, whereas our proposed framework explicitly addresses these gaps by simultaneously modeling modality-specific structural information and cross-modal semantic alignment within a unified, interpretable representation learning framework.

We propose the \textbf{Structural-Semantic Unifier (SSU)}, a novel graph-based framework explicitly designed to unify modality-specific structural dependencies and cross-modal semantic alignment for multimodal sentiment analysis. SSU is built on the fundamental insight that effective multimodal fusion must simultaneously preserve the internal structures inherent in each modality and resolve semantic disparities across modalities—particularly crucial for capturing nuanced affective signals that may be asynchronous or ambiguous.

SSU constructs modality-specific graphs to preserve fine-grained structural dependencies—using syntactic parsing for text and text-guided attention for audio and visual modalities—while introducing semantic anchor as shared reference nodes grounded in linguistic context. These anchors serve to unify heterogeneous modality graphs, enabling coherent and interpretable cross-modal alignment. To further enhance representation quality, SSU incorporates a multiview contrastive learning objective that jointly enforces task-specific discrimination, structural consistency, and semantic coherence. Together, these components yield robust, discriminative, and semantically aligned multimodal representations.

The main contributions are summarized as follows:
\begin{itemize}
\item We introduce \textbf{SSU}, a unified multimodal fusion framework designed to integrate modality-specific structural dependencies with semantic alignment across modalities.

\item We develop a novel \textbf{modality-specific graph construction} method, leveraging syntactic parsing for textual structure and text-guided attention for audio-visual semantic relevance, further strengthened by semantic anchor to ensure coherent cross-modal alignment.

\item We propose a comprehensive \textbf{multiview contrastive learning} paradigm that simultaneously enforces structural consistency, semantic coherence, and task-level discriminability across heterogeneous modality representations.

\item Extensive evaluations on widely-adopted benchmarks, including CMU-MOSI and CMU-MOSEI, demonstrate that SSU significantly outperforms existing state-of-the-art approaches in both predictive performance and computational efficiency.
\end{itemize}

\section{Related Work}
Multimodal Sentiment Analysis (MSA) aims to understand human affective states by jointly analyzing textual, visual, and acoustic signals. Early works typically employ unimodal encoders coupled with straightforward multimodal fusion strategies, such as concatenation~\cite{zadeh2017tensor, wu2023multimodal} and attention-based integration~\cite{tsai2019multimodal, hazarika2020misa, yu2021learning}. Although these methods achieve reasonable results, they often treat each modality independently, neglecting explicit structural relationships among modalities. Consequently, these approaches may fail to adequately capture nuanced multimodal interactions, especially in complex linguistic scenarios involving sarcasm, irony, or negation.

To address these limitations, recent studies have explored advanced encoders and refined learning objectives. UniMSE~\cite{hu2022unimse} leverages a pre-trained T5 model for textual representations, while audio and visual modalities are encoded separately through LSTMs, combined with cross-modal and inter-sample contrastive learning to enhance feature interaction. SeMuL-PCD~\cite{anand2023multi} integrates modality-specific knowledge via cross-modal peer distillation and contrastive objectives, facilitating richer multimodal representations. MMIM~\cite{han2021improving} preserves discriminative multimodal information through mutual information maximization between unimodal and joint representations. Other related efforts include handling modality incompleteness~\cite{peng2021adaptive, lin2023missmodal, li2024unified}, dynamic attention modulation strategies~\cite{su2020msaf, chen2022weighted}, recurrent modeling of conversational sentiment dynamics~\cite{huddar2021attention}, and language-guided multimodal interactions~\cite{mai2021analyzing}.

In parallel, graph-based methods have demonstrated potential in explicitly modeling structural dependencies among modalities. MMGCN~\cite{hu2021mmgcn} introduces multimodal graph convolution networks to jointly capture intra- and inter-modal dependencies, while MGNNS~\cite{yang2021multimodal} constructs sentiment-aware graphs per modality to extract global sentiment characteristics. AMGIN~\cite{gong2024adaptive} employs adaptive gating mechanisms to dynamically integrate modality-specific graphs, promoting robustness and modality interaction. GraphMFT~\cite{li2023graphmft} formulates conversation-based emotion recognition as a heterogeneous graph modeling problem, effectively capturing fine-grained contextual interactions within and across modalities.

Despite their effectiveness, existing graph-based methods typically employ static or modality-isolated graphs, lacking a unified mechanism for semantic alignment. To address this, we propose dynamically constructed modality-specific graphs guided by syntactic parsing and text-guided attention, integrated through semantic anchor and enhanced by multiview contrastive learning.

\section{Methodology}
\begin{figure*}[t!]
\centering
\includegraphics[width=1\linewidth]{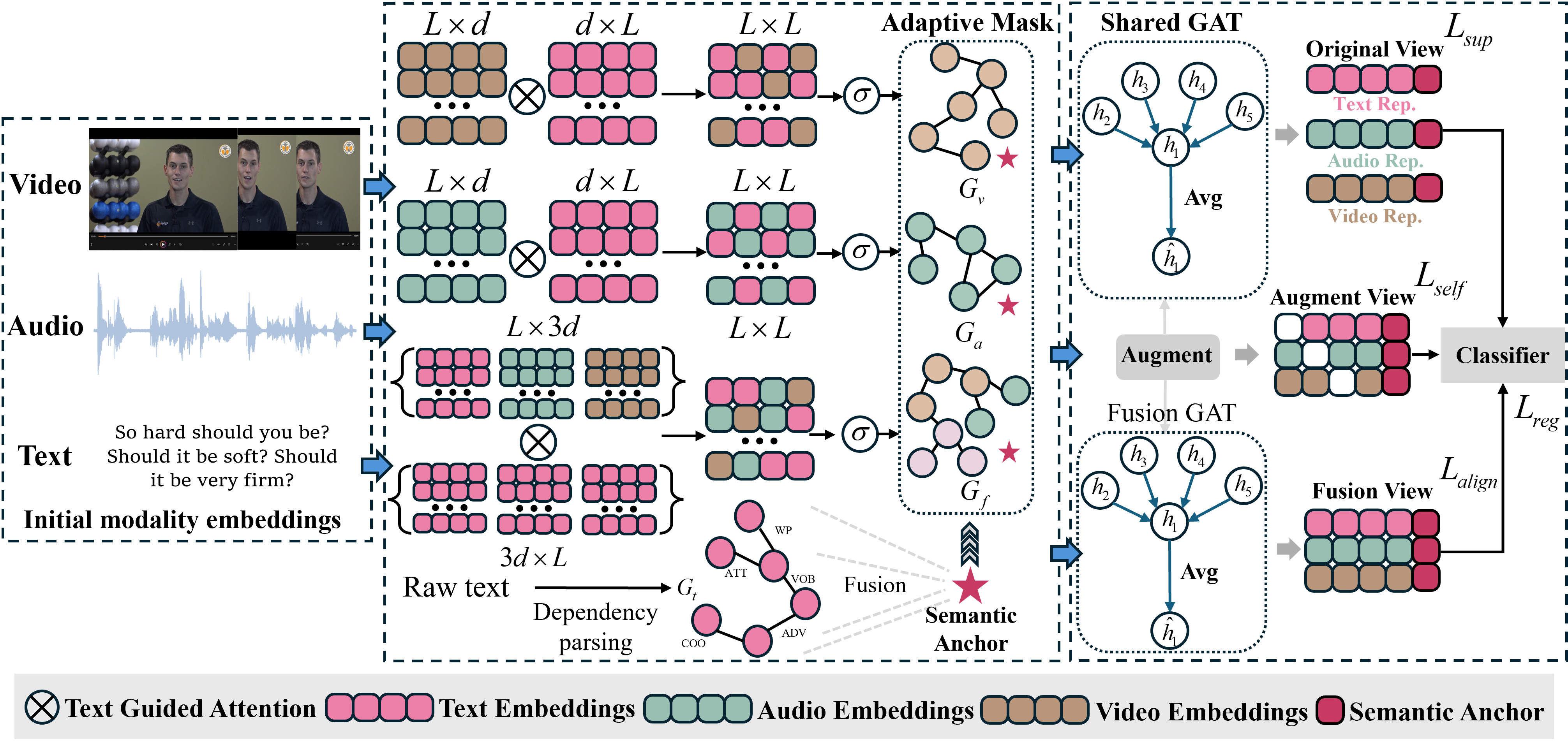}
\caption{Overview of the SSU framework. Modality-specific graphs are constructed via syntactic parsing for text and text-guided attention for audio and video, capturing intra-modal \textbf{structural dependencies}. \textbf{Semantic anchor} derived from global text semantics is injected into all graphs to unify modalities. A shared GAT encodes structure-aware representations, while a fusion GAT integrates anchor-enhanced graphs. Multi-view contrastive learning is applied across original, augmented, and anchor-guided views to enforce structural consistency and semantic alignment.}
\label{fig:framework}
\end{figure*}

We propose the \textbf{Structural-Semantic Unifier (SSU)}, a unified framework that integrates intra-modal structure and cross-modal semantics for multimodal sentiment analysis. As illustrated in Figure~\ref{fig:framework}, SSU comprises three components: (1) modality-specific graph construction to capture structural dependencies; (2) semantic anchor to align modality graphs via language-guided reference nodes; and (3) a multi-view contrastive objective that enforces structural consistency, semantic coherence, and task-level discrimination.

\subsection{Modality-Specific Graph Construction}
Our approach constructs modality-specific graphs to explicitly model inherent structural dependencies and semantic alignments. This process is a crucial departure from conventional fusion techniques, laying the foundation for a richer, more expressive representation space. The methodology is executed in two primary stages: initial feature enhancement and dynamic graph generation.

The textual modality provides an explicit semantic structure crucial for sentiment analysis. We construct the text graph, $\mathcal{G}^t$, based on the syntactic dependency parse tree of the textual transcript. Each word corresponds to a node in the graph, and directed edges represent grammatical relationships (e.g., subject, object, or modifiers). This graph serves as a core structural element, preserving the fundamental linguistic organization that guides the processing of other modalities. 

We first enhance the unimodal representations by incorporating semantic information from the text. This is achieved via a cross-attention mechanism, where the textual representations serve as the queries to enrich the non-textual representations. Given the initial feature sequences $X^t\in \mathcal{R}^{T_t\times d}$ and $X^m\in \mathcal{R}^{T_m\times d}$ for text and modality $m\in \{a,v\}$ respectively, the augmented feature sequence $\hat{H}^m$ is obtained by Eq~\ref{eq:ematt}.
\begin{equation}
    \hat{H}^m = \mathrm{Attention}(X^t, X^m).
    \label{eq:ematt}
\end{equation}
The final enhanced representation, $H'^m$, is then derived by adding this augmented feature back to the original feature sequence via a residual connection (Eq.~\ref{eq:resiadd}).
\begin{equation}
    H'^m=X^m+\hat{H}^m.
    \label{eq:resiadd}
\end{equation}

For the audio and visual modalities, we dynamically construct a sparse and robust graph based on the cross-modal semantic relevance. We compute a raw score matrix $S^m\in \mathcal{R}^{T_m\times T_t}$ that quantifies the semantic similarity between segments of modality $m$ and the text by Eq.~\ref{eq:simsem}.
\begin{equation}
    S^m=\frac{(X^m \cdot W_Q)(X^t\cdot W_K)^\top}{\sqrt{d'}}+B_{pos},
    \label{eq:simsem}
\end{equation}
where $W_Q, W_K\in\mathcal{R}^{d\times d'}$ are learnable projection matrices, and $B_{pos}$ is a learnable temporal position bias matrix that encourages connections between temporally proximate segments. The raw scores are then normalized via a softmax function to obtain the base adjacency matrix $A'^m$, which is subsequently symmetrized to ensure bidirectional relationships (Eq.~\ref{eq:bsadj}).
\begin{equation}
    A'^m=\frac{1}{2}(\mathrm{softmax}(S^m)+\mathrm{softmax}(S^m)^\top).
    \label{eq:bsadj}
\end{equation}

To enhance robustness and computational efficiency, we introduce an adaptive sparsification strategy that prunes noisy connections. For each sample, we compute a dynamic threshold $\tau_m$ based on the mean ($\mu_m$) and deviation ($\sigma_m$) of the base matrix $A'^m$, $\tau_m=\mu_m+\lambda \sigma_m$, where $\lambda$ is a learnable scaling factor. We then generate a binary mask matrix $M^m$ that only retains edges with scores exceeding this threshold, $M^{m_{ij}}=\mathbb{I}(A'^{m_{ij}}) \ge \tau_m$, where $\mathbb{I}(\cdot)$ is the indicator function. The final, dynamically pruned adjacency matrix $A'^m$ is obtained by applying this mask to the base matrix, as Eq.~\ref{eq:adjcal}.
\begin{equation}
    A^m = A'^m\odot M^m.
    \label{eq:adjcal}
\end{equation}

\begin{figure}[t!]
    \centering
    \includegraphics[width=1\linewidth]{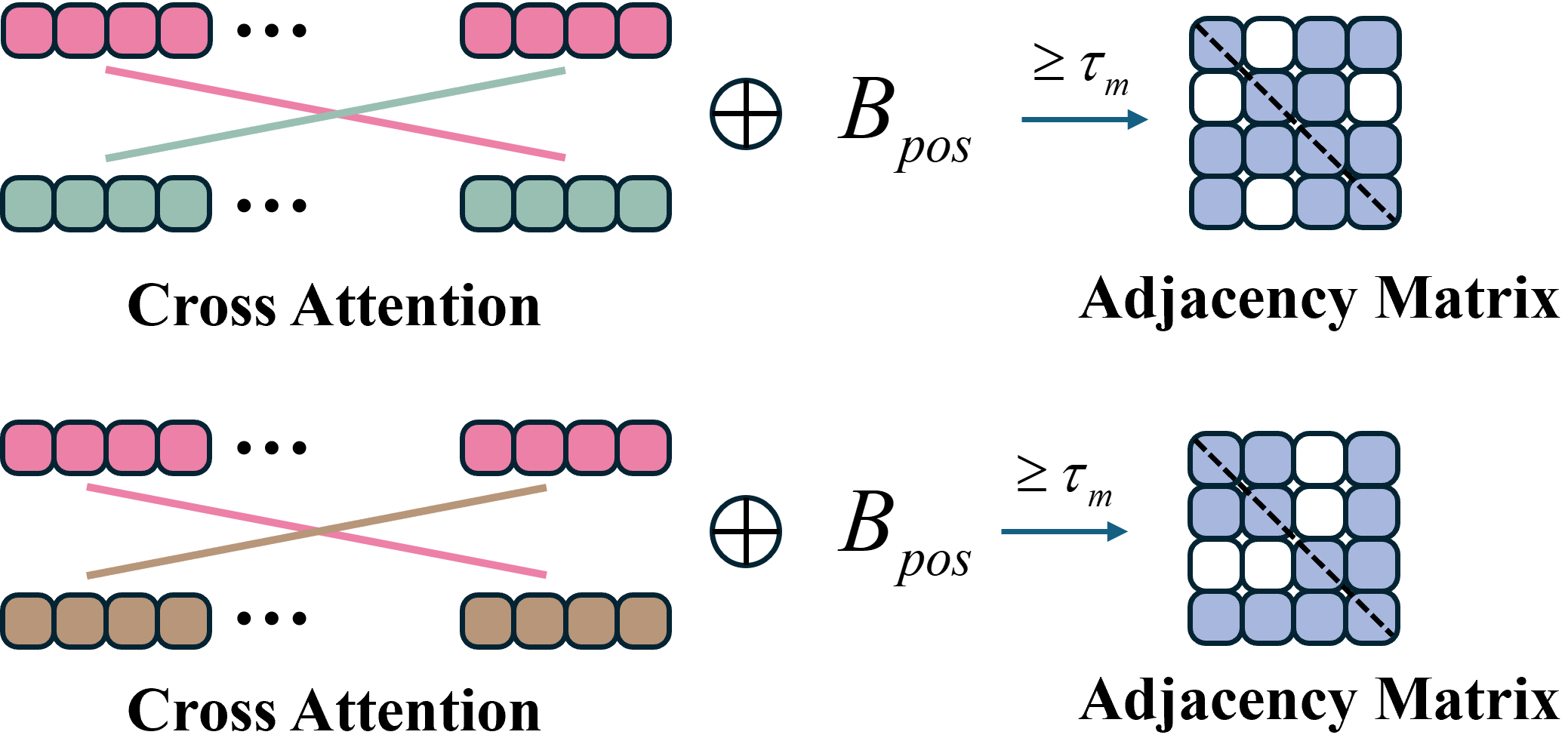}
    \caption{Illustration of text-guided graph construction for audio and visual modalities.}
    \label{fig:textguidedattention}
\end{figure}

To better illustrate this process, Figure~\ref{fig:textguidedattention} provides a visual explanation of how our text-guided attention mechanism enables semantic graph construction. Specifically, textual tokens are first used to query the audio and visual segments through a cross-modal attention operation. Each non-text segment obtains a text-informed attention vector. Next, pairwise similarity between attention vectors within each modality is computed to capture the semantic relatedness of non-textual segments under textual guidance. The resulting affinity matrix forms the basis of the graph structure, which is then sparsified by thresholding, as described above.
\subsection{Semantic Anchor Integration}
While modality-specific graphs capture intra-modal structural dependencies, they still operate within their individual representational spaces, potentially leading to semantic misalignment across modalities. To address this issue, we introduce \textbf{semantic anchor}—shared latent nodes derived from global text semantics—to bridge heterogeneous modality graphs and enforce cross-modal consistency. The semantic anchor is instantiated as a global representation vector $z_a \in \mathbb{R}^{d}$, computed via average pooling over the contextualized text sequence:
\begin{equation}
    z_a = \mathrm{AvgPool}(X^t).
\end{equation}
This anchor captures the utterance-level semantic intent expressed in the textual modality, which often serves as the most reliable source of sentiment cues. Unlike individual text tokens, the anchor encodes a holistic semantic abstraction.

To unify modalities structurally and semantically, we inject the semantic anchor into each modality-specific graph $\mathcal{G}^m = (\mathcal{V}^m, \mathcal{E}^m)$ by:
\begin{itemize}
    \item Adding the anchor node to the node set: $\mathcal{V}^m \leftarrow \mathcal{V}^m \cup \{z_a\}$;
    \item Connecting the anchor to all modality nodes via attention-based edge weights:
    \begin{equation}
        \beta_i^m = \mathrm{softmax}(z_a^\top W_a H'^m_i),
    \end{equation}
    where $W_a \in \mathbb{R}^{d \times d}$ is a learnable projection matrix, and $\beta_i^m$ indicates the strength of semantic relevance between $z_a$ and node $i$ in modality $m$.
\end{itemize}
The resulting augmented graph contains a shared semantic hub that aligns structurally disjoint modality graphs into a common space.

We further construct a \textbf{fusion graph} $\mathcal{G}^f$ that combines all modality nodes and the semantic anchor. This graph allows interactions not only within each modality but also across modalities via anchor-mediated connections. Specifically, edges are retained between:
\begin{itemize}
    \item Modality-specific nodes (from $\mathcal{G}^a$, $\mathcal{G}^v$, and $\mathcal{G}^t$),
    \item Each node and the anchor node,
    \item Select cross-modality pairs with high semantic affinity to the anchor.
\end{itemize}

This design allows the anchor to serve as both a semantic aggregator and a structural bridge across modalities, effectively promoting coherent and interpretable fusion.

\subsection{Multi-View Contrastive Learning Objective}
To further enhance the discriminability, robustness, and semantic coherence of the learned representations, we incorporate a \textbf{multi-view contrastive learning objective} into the SSU framework. This design encourages alignment across heterogeneous modalities, resilience to structural perturbations, and preservation of task-relevant semantics.

We formulate three distinct views of each input:
\begin{itemize}
    \item \textbf{Original View} (\textit{structure-aware}): modality-specific graphs $\mathcal{G}^m$ are encoded via a shared GAT to capture intra-modal structure and anchor-enhanced semantics.
    \item \textbf{Fusion View} (\textit{anchor-guided}): modality graphs are unified through the semantic anchor into a fusion graph $\mathcal{G}^f$, which is encoded using a separate GAT to model cross-modal alignment.
    \item \textbf{Augmented View} (\textit{noisy variant}): random perturbations (e.g., edge addition/deletion) are applied to $\mathcal{G}^m$ to simulate noisy conditions, and representations are re-encoded using the same shared GAT.
\end{itemize}

Let $z_{\text{ori}}$, $z_{\text{aug}}$, and $z_{\text{fuse}}$ denote the representations derived from the original, augmented, and fusion views, respectively. Our contrastive objective consists of three components:

\paragraph{Supervised Discrimination Loss.}
Given labeled training samples, we enforce supervised separation via a standard cross-entropy loss:
\begin{equation}
    \mathcal{L}_{\text{sup}} = \mathrm{CE}(\mathrm{Classifier}(z_{\text{ori}}), y).
\end{equation}

\paragraph{Structural Consistency Loss.}
To ensure robustness to structural perturbations, we apply a self-supervised contrastive loss between $z_{\text{ori}}$ and $z_{\text{aug}}$:
\begin{equation}
    \mathcal{L}_{\text{self}} = -\log \frac{\exp(\mathrm{sim}(z_{\text{ori}}, z_{\text{aug}})/\eta)}{\sum\limits_{z^- \in \mathcal{N}} \exp(\mathrm{sim}(z_{\text{ori}}, z^-)/\eta)},
\end{equation}
where $\eta$ is a temperature parameter and $\mathcal{N}$ denotes the set of negative samples in the mini-batch.

\paragraph{Semantic Alignment Loss.}
To unify multimodal semantics, we encourage the anchor-guided fusion representation to align with the original:
\begin{equation}
    \mathcal{L}_{\text{align}} = \| z_{\text{fuse}} - z_{\text{ori}} \|_2^2.
\end{equation}

\paragraph{Final Objective.}
The total training loss is a weighted sum of the three objectives:
\begin{equation}
    \mathcal{L} = \mathcal{L}_{\text{sup}} + \lambda_{\text{self}} \mathcal{L}_{\text{self}} + \lambda_{\text{align}} \mathcal{L}_{\text{align}},
\end{equation}
where $\lambda_{\text{self}}$ and $\lambda_{\text{align}}$ are hyperparameters that control the balance among different learning signals.

This multi-view contrastive formulation encourages the model to learn structure-aware, semantically aligned, and perturbation-resilient representations—crucial for robust multimodal sentiment understanding.

\section{Experiments}
\begin{table*}[t!]
\centering
\resizebox{\textwidth}{!}{
\begin{tabular}{l|cccc|cccc}
\toprule
\multirow{2}{*}{\textbf{Method}} & \multicolumn{4}{c|}{\textbf{CMU-MOSI}} & \multicolumn{4}{c}{\textbf{CMU-MOSEI}} \\
& ACC2$\uparrow$ & F1$\uparrow$ & ACC7$\uparrow$ & MAE$\downarrow$ & ACC2$\uparrow$ & F1$\uparrow$ & ACC7$\uparrow$ & MAE$\downarrow$ \\
\midrule
\multicolumn{9}{c}{\textit{Attention-based Methods}} \\
\midrule
CIA~\cite{chauhan2019context} & 79.88\% & 79.54\% & 38.92\% & 0.9147 & 80.37\% & 78.23\% & 50.14\% & 0.6835 \\
MAT~\cite{delbrouck2020modulated} & 80.00\% & 80.00\% & 35.41\% & 0.9230 & 82.00\% & 82.00\% & 47.32\% & 0.7067 \\
TBJE~\cite{delbrouck2020transformer} & 81.00\% & 78.00\% & 40.00\% & 0.8950 & 82.48\% & 65.54\% & 45.52\% & 0.7441 \\
GATE~\cite{kumar2020gated} & 83.91\% & 81.17\% & 42.85\% & 0.8600 & 85.27\% & 84.08\% & 53.26\% & 0.6200 \\
MPT~\cite{cheng2021multimodal} & 82.80\% & 82.90\% & 43.20\% & 0.7900 & 82.60\% & 82.80\% & 50.60\% & 0.5800 \\
UniMSE~\cite{hu2022unimse} & 86.90\% & 86.42\% & 48.68\% & 0.6914 & 87.50\% & 87.46\% & 54.39\% & 0.5238 \\
SPECTRA~\cite{yu2023speech} & 87.50\% & 87.20\% & 49.20\% & 0.6600 & 87.34\% & 87.10\% & 53.95\% & 0.5350 \\
MMML+FusionNet~\cite{wu2023multimodal} & 88.16\% & 88.15\% & 48.25\% & 0.6429 & 86.73\% & 86.49\% & 51.54\% & 0.5154 \\
CMPT~\cite{reza2025robust} & 88.51\% & 88.34\% & 49.13\% & 0.6223 & 87.03\% & 86.92\% & 53.64\% & 0.5271 \\
\midrule
\multicolumn{9}{c}{\textit{Graph-based Methods}} \\
\midrule
MMGraph~\cite{mai2020analyzing} & 81.40\% & 81.70\% & 49.70\% & 0.6082 & 80.60\% & 80.50\% & 32.10\% & 0.9331 \\
GraphCAGE~\cite{wu2021graph} & 82.10\% & 82.10\% & 35.40\% & 0.9333 & 81.70\% & 81.80\% & 48.90\% & 0.6092 \\
CJTF-BERT~\cite{lu2024coordinated} & 86.50\% & 86.40\% & 47.00\% & 0.7046 & 86.10\% & 86.04\% & 52.90\% & 0.5137 \\
MoSARe~\cite{moradinasab2025towards} & 88.37\% & 88.10\% & 49.21\% & 0.6346 & 87.03\% & 86.91\% & 53.90\% & 0.5223 \\
\midrule
 \textbf{SSU (Ours)} & \textbf{89.32\%} & \textbf{89.28\%} & \textbf{51.89\%} & \textbf{0.5666} & \textbf{87.93\%} & \textbf{87.72\%} & \textbf{55.29\%} & \textbf{0.5090} \\
\bottomrule
\end{tabular}
}
\caption{Comparison with state-of-the-art models on CMU-MOSI and CMU-MOSEI datasets. Metrics include binary accuracy (ACC2$\uparrow$), F1-score (F1$\uparrow$), 7-class accuracy (ACC7$\uparrow$), and mean absolute error (MAE$\downarrow$), where $\uparrow$ indicates higher is better and $\downarrow$ indicates lower is better.}
\label{tab:sota-multimodal}
\end{table*}

\begin{table}[t!]
    \centering
    \setlength{\tabcolsep}{1.1mm}
    \begin{tabular}{l|cccc}
        \toprule
        \multicolumn{5}{c}{\textbf{CMU-MOSEI}} \\
        \midrule
        \textbf{Model} & ACC2$\uparrow$ & F1$\uparrow$ & ACC7$\uparrow$ & MAE$\downarrow$ \\
        \midrule
        LLaMA-3.1-8B & 86.72\% & 86.62\% & 53.30\% & 0.52 \\
        Qwen-2.5-7B & 86.55\% & 86.47\% & 53.12\% & 0.52 \\
        Mistral-7B & 85.33\% & 85.20\% & 52.84\% & 0.56 \\
        Gemma-2B & 84.15\% & 84.05\% & 51.35\% & 0.59 \\
        
        \textbf{SSU (Ours)} & \textbf{87.93\%} & \textbf{87.72\%} & \textbf{55.29\%} & \textbf{0.51} \\
        \toprule
        \multicolumn{5}{c}{\textbf{CMU-MOSI}} \\
        \midrule
        \textbf{Model} & ACC2$\uparrow$ & F1$\uparrow$ & ACC7$\uparrow$ & MAE$\downarrow$ \\
        \midrule
        LLaMA-3.1-8B & 87.65\% & 87.55\% & 44.30\% & 0.72 \\
        Qwen-2.5-7B & 87.43\% & 87.30\% & 44.00\% & 0.73 \\
        Mistral-7B & 86.21\% & 86.12\% & 43.55\% & 0.75 \\
        Gemma-2B & 84.73\% & 84.65\% & 41.90\% & 0.80 \\
        
        \textbf{SSU (Ours)} & \textbf{89.32\%} & \textbf{89.28\%} & \textbf{51.89\%} & \textbf{0.57} \\
        \bottomrule
    \end{tabular}
    \caption{
    Comparison of SSU and LLM-based graph constructors on CMU-MOSEI and CMU-MOSI. Metrics include accuracy (ACC2, ACC7), F1, and MAE. SSU outperforms all LLMs while incurring minimal runtime cost.
    }
    \label{tab:llm_graph_compare}
\end{table}

\subsection{Datasets and Metrics}
We evaluate our model on two standard multimodal sentiment analysis benchmarks: \textbf{CMU-MOSI}~\cite{zadeh2016mosi} and \textbf{CMU-MOSEI}~\cite{zadeh2018multimodal}. 

CMU-MOSI consists of 2,199 opinionated utterances from 93 video reviews, with sentiment scores ranging from $-3$ to $+3$. CMU-MOSEI expands this to 23,453 utterances from over 1,000 speakers. We follow standard splits for training, validation, and testing. To comprehensively evaluate model performance, we adopt four metrics:
\begin{itemize}
    \item \textbf{ACC2}: Binary accuracy (positive vs. negative sentiment).
    \item \textbf{F1-score}: Harmonic mean of precision and recall.
    \item \textbf{ACC7}: 7-class accuracy based on discretized sentiment scores.
    \item \textbf{MAE}: Mean Absolute Error for continuous sentiment regression.
\end{itemize}
\begin{figure}[t!]
    \centering
    \includegraphics[width=1\linewidth]{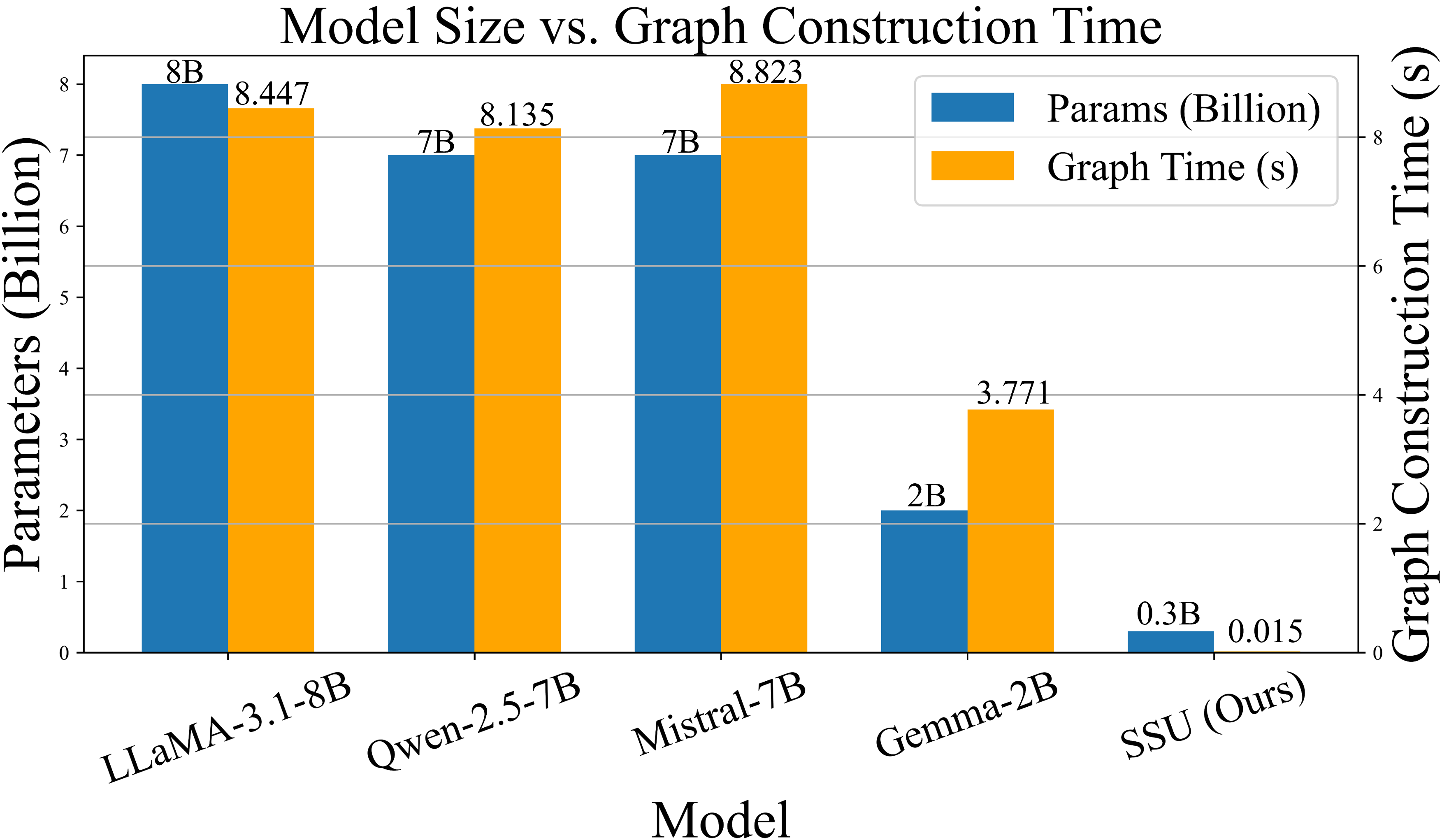}
    \caption{
    Comparison of model size and graph construction time. SSU achieves the best efficiency–performance trade-off, requiring only 0.3B parameters and approximately 0.015s per batch, significantly outperforming LLM-based graph constructors with substantially lower computational overhead.
    }
    \label{fig:llmvssu}
\end{figure}

All experiments are conducted on a workstation with 8 NVIDIA A100 GPUs and 512 GB RAM, using PyTorch 1.8.2 (CUDA 11.1) and SpaCy 3.5.0. We fix the random seed to 68 for reproducibility. The model is trained with a batch size of 128, sequence length of 128, hidden size of 128, and learning rate of $1\text{e}-5$. We apply lightweight data augmentation and contrastive learning to enhance structural and semantic alignment.
\begin{figure}[t!]
    \centering
    \includegraphics[width=1\linewidth]{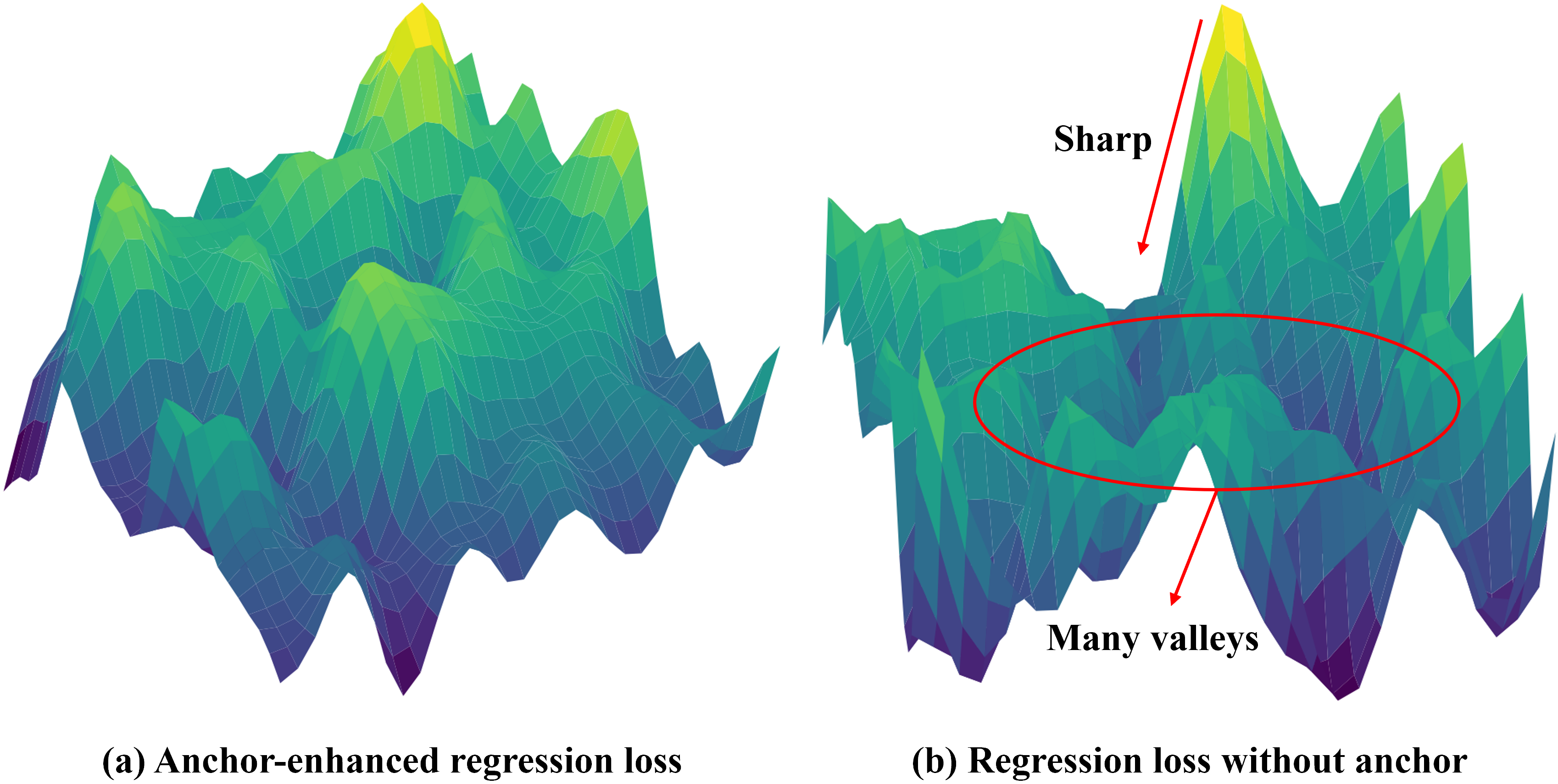}
    \caption{Loss surface visualization with (left) and without (right) semantic anchor. The anchor yields a smoother landscape, indicating improved optimization stability.
    }
    \label{fig:anchor_loss_surface}
\end{figure}
\begin{table}[t!]
    \centering
    \setlength{\tabcolsep}{1mm}
    \begin{tabular}{l|cccc}
        \toprule
        \multicolumn{5}{c}{\textbf{CMU-MOSI}} \\
        \midrule
        Method & ACC2$\uparrow$ & F1$\uparrow$ & ACC7$\uparrow$ & MAE$\downarrow$ \\
        \midrule
        w/o Semantic Anchor & 86.89\% & 86.86\% & 48.54\% & 0.5990 \\
        
        \textbf{Full model (SSU)} & \textbf{89.32\%} & \textbf{89.28\%} & \textbf{51.89\%} & \textbf{0.5666} \\
        \toprule
        \multicolumn{5}{c}{\textbf{CMU-MOSEI}} \\
        \midrule
        Method & ACC2$\uparrow$ & F1$\uparrow$ & ACC7$\uparrow$ & MAE$\downarrow$ \\
        \midrule
        w/o Semantic Anchor & 87.37\% & 87.27\% & 53.94\% & 0.5175 \\
        
        \textbf{Full model (SSU)} & \textbf{87.93\%} & \textbf{87.72\%} & \textbf{55.29\%} & \textbf{0.5090} \\
        \bottomrule
        \end{tabular}
        \caption{Effect of semantic anchor. Performance drops without the anchor confirm its importance in semantic alignment and stable graph learning.}
    \label{tab:ablation_anchor}
\end{table}
\subsection{Comparison with SOTA Methods}
We compare SSU with recent state-of-the-art methods across both attention-based and graph-based multimodal sentiment models.

\textbf{Attention-based models} (e.g., CIA~\cite{chauhan2019context}, GATE~\cite{kumar2020gated}, MPT~\cite{cheng2021multimodal}, UniMSE~\cite{hu2022unimse}, SPECTRA~\cite{yu2023speech}) focus on dynamic cross-modal alignment using attention or contrastive objectives. \textbf{Graph-based methods} (e.g., MMGraph~\cite{mai2020analyzing}, CJTF-BERT~\cite{lu2024coordinated}, MoSARe~\cite{moradinasab2025towards}) explicitly model structural dependencies, often relying on static graphs.

As shown in Table~\ref{tab:sota-multimodal}, SSU achieves new state-of-the-art results on both CMU-MOSI and CMU-MOSEI across all metrics. Compared to the graph-based baseline MoSARe, SSU improves ACC2 by $+1.6\%$ on MOSI and $+1.3\%$ on MOSEI, while reducing MAE and improving ACC7.

\subsection{SSU versus LLM-based Graph Constructors}
We compare SSU with representative large language models (LLMs) that construct modality graphs via offline prompting~\cite{touvron2023llama, bai2025qwen2, hamzah2024multimodal, team2024gemma}. While LLMs encode strong linguistic priors, they suffer from high graph construction latency and require multi-billion parameter models, limiting their practicality for real-time applications.

In contrast, SSU constructs modality-specific graphs online and without supervision. For fair comparison, all LLMs are 8-bit quantized and receive concatenated text, audio, and visual embeddings. Hidden states from their penultimate layers are linearly projected to form adjacency matrices. SSU completes this process in just \textbf{0.015s}, significantly faster than LLaMA-3.1-8B (8.45s), Qwen-2.5-7B (8.14s), Mistral-7B (8.82s), and Gemma-2B (3.77s), as shown in Figure~\ref{fig:llmvssu}. Despite having only \textbf{0.3B} parameters, SSU consistently outperforms these LLMs on CMU-MOSI and CMU-MOSEI benchmarks; see Table~\ref{tab:llm_graph_compare}.

\begin{figure*}[t!]
    \centering
    \includegraphics[width=\linewidth]{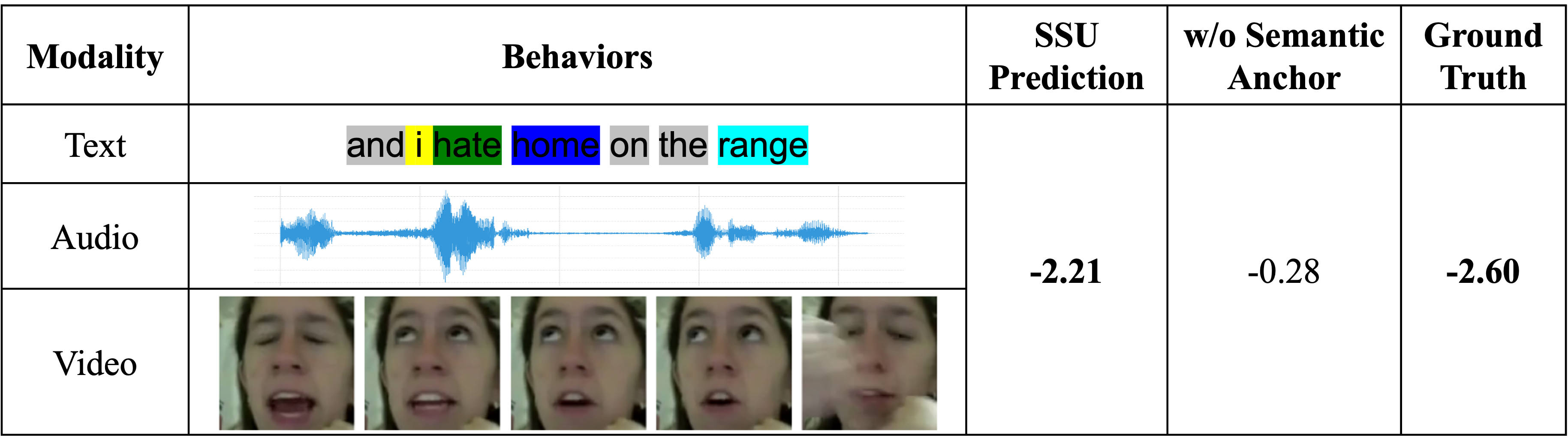}
    \caption{
    Prediction comparison on a negative utterance. Anchors enhance modality alignment and semantic consistency, leading to better sentiment estimation. 
    Predictions are raw regression outputs before activation (i.e., pre-logit values).
    }
    \label{fig:casestudy}
\end{figure*}
\subsection{Effect of Semantic Anchor}
We assess the impact of the semantic anchor via ablation. As shown in Table~\ref{tab:ablation_anchor}, its removal causes a notable performance drop, including a $2.42\%$ F1 decrease and higher MAE, indicating weaker semantic discrimination.

Figure~\ref{fig:anchor_loss_surface} visualizes the loss surface, where the anchor leads to a smoother, more convex landscape, while its absence results in sharper, unstable minima. This highlights the anchor’s role in stabilizing training and enhancing cross-modal semantic alignment.

\begin{table}[t!]
    \centering
    \setlength{\tabcolsep}{1.1mm}
    \begin{tabular}{l|cccc}
        \toprule
        \textbf{Loss Combination} & \textbf{ACC2}$\uparrow$ & \textbf{F1}$\uparrow$ & \textbf{ACC7}$\uparrow$ & \textbf{MAE}$\downarrow$ \\
        \midrule
        \multicolumn{5}{c}{\textit{CMU-MOSI}} \\
        \midrule
        $\mathcal{L}_{\text{reg}}$ only & 86.45\% & 86.38\% & 46.30\% & 0.7031 \\
        $\mathcal{L}_{\text{reg}}$ + $\mathcal{L}_{\text{align}}$ & 87.32\% & 87.25\% & 47.40\% & 0.6837 \\
        $\mathcal{L}_{\text{reg}}$ + $\mathcal{L}_{\text{self}}$ & 88.10\% & 88.00\% & 48.55\% & 0.6622 \\
        $\mathcal{L}_{\text{reg}}$ + $\mathcal{L}_{\text{sup}}$ & 88.54\% & 88.46\% & 49.30\% & 0.6455 \\
        $\mathcal{L}_{\text{reg}}$ + $\mathcal{L}_{\text{self}}$ + $\mathcal{L}_{\text{align}}$ & 88.73\% & 88.65\% & 50.10\% & 0.6233 \\
        $\mathcal{L}_{\text{reg}}$ + $\mathcal{L}_{\text{sup}}$ + $\mathcal{L}_{\text{align}}$ & 88.91\% & 88.84\% & 50.66\% & 0.6074 \\
        $\mathcal{L}_{\text{reg}}$ + $\mathcal{L}_{\text{sup}}$ + $\mathcal{L}_{\text{self}}$ & 88.97\% & 88.90\% & 51.02\% & 0.5901 \\
        
        \textbf{Full (Ours)} & \textbf{89.32\%} & \textbf{89.28\%} & \textbf{51.89\%} & \textbf{0.5666} \\
        \midrule
        \multicolumn{5}{c}{\textit{CMU-MOSEI}} \\
        \midrule
        $\mathcal{L}_{\text{reg}}$ only & 86.15\% & 86.08\% & 52.13\% & 0.6034 \\
        $\mathcal{L}_{\text{reg}}$ + $\mathcal{L}_{\text{align}}$ & 86.94\% & 86.90\% & 53.21\% & 0.5893 \\
        $\mathcal{L}_{\text{reg}}$ + $\mathcal{L}_{\text{self}}$ & 87.22\% & 87.16\% & 54.00\% & 0.5756 \\
        $\mathcal{L}_{\text{reg}}$ + $\mathcal{L}_{\text{sup}}$ & 87.47\% & 87.41\% & 54.31\% & 0.5632 \\
        $\mathcal{L}_{\text{reg}}$ + $\mathcal{L}_{\text{self}}$ + $\mathcal{L}_{\text{align}}$ & 87.61\% & 87.55\% & 54.65\% & 0.5528 \\
        $\mathcal{L}_{\text{reg}}$ + $\mathcal{L}_{\text{sup}}$ + $\mathcal{L}_{\text{align}}$ & 87.74\% & 87.68\% & 54.91\% & 0.5392 \\
        $\mathcal{L}_{\text{reg}}$ + $\mathcal{L}_{\text{sup}}$ + $\mathcal{L}_{\text{self}}$ & 87.83\% & 87.71\% & 55.13\% & 0.5264 \\
        
        \textbf{Full (Ours)} & \textbf{87.93\%} & \textbf{87.72\%} & \textbf{55.29\%} & \textbf{0.5090} \\
        \bottomrule
    \end{tabular}
    \caption{
    Ablation study on the multi-view contrastive objective. The full model combines all three contrastive losses and the main regression loss.}
    \label{tab:loss-ablation}
\end{table}
\begin{figure}[t!]
    \centering
    \includegraphics[width=1\linewidth]{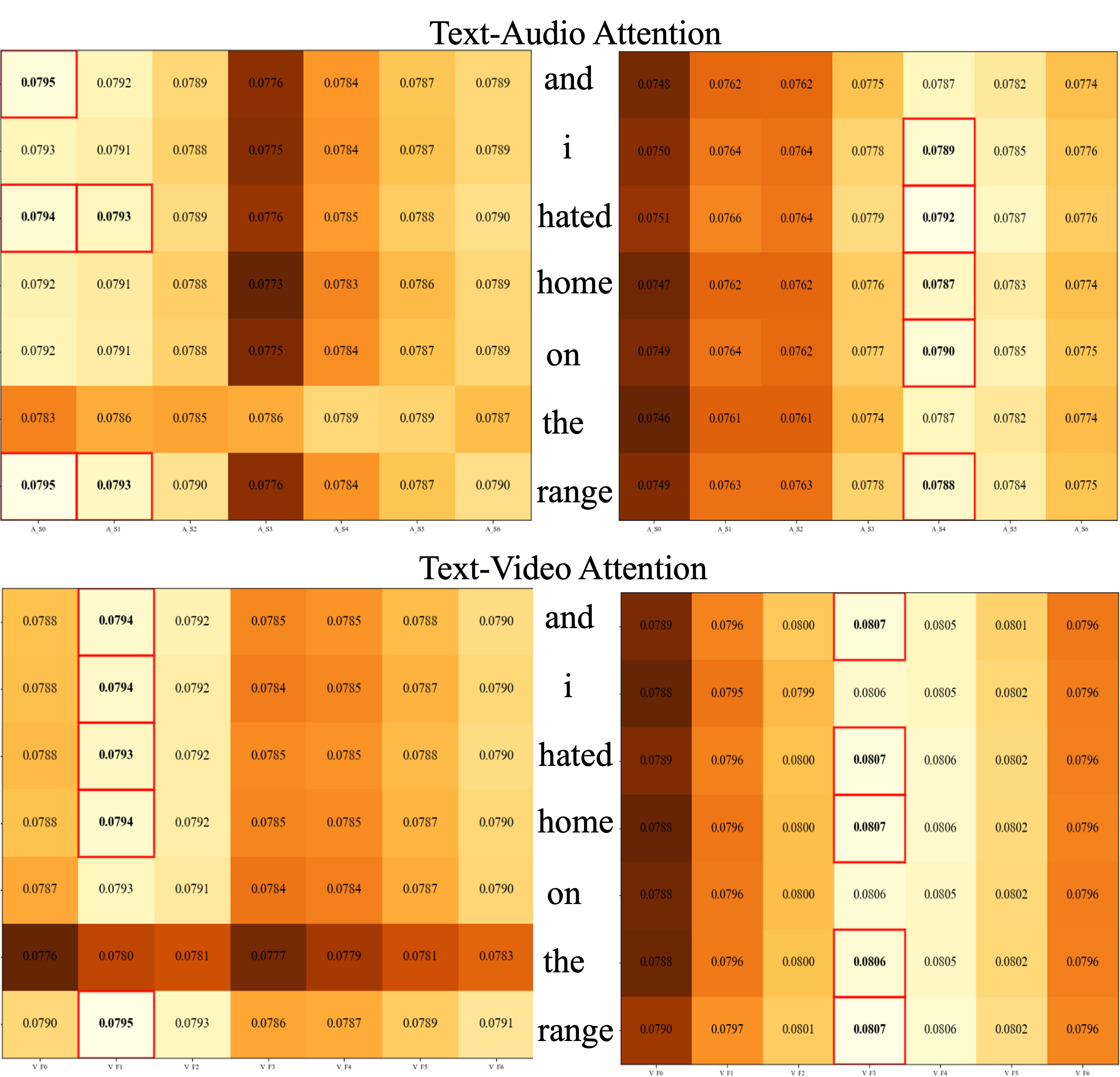}
    \caption{Cross-modal attention maps with (right) and without (left) semantic anchor. Anchors produce more concentrated and semantically meaningful attention.}
    \label{fig:attentionmap}
\end{figure}
\subsection{Analysis of Multiview Contrastive Objective}
To evaluate the impact of each component in our multiview contrastive framework, we conduct an ablation study by progressively adding the contrastive losses to the main regression objective. As shown in Table~\ref{tab:loss-ablation}, using only $\mathcal{L}_{\text{reg}}$ already yields competitive results, demonstrating the strength of our SSU design. However, performance steadily improves as different contrastive signals ($\mathcal{L}_{\text{align}}$, $\mathcal{L}_{\text{sup}}$, and $\mathcal{L}_{\text{self}}$) are incorporated.

Notably, combining all three contrastive objectives results in consistent gains across both CMU-MOSI and CMU-MOSEI, with improvements in ACC2, F1, and ACC7, and reductions in MAE. These results confirm that each loss contributes complementary supervision: $\mathcal{L}_{\text{align}}$ promotes cross-modal consistency, $\mathcal{L}_{\text{sup}}$ enhances semantic discriminability, and $\mathcal{L}_{\text{self}}$ improves intra-modal coherence. Together, they encourage the model to learn more robust and semantically aligned representations.

\subsection{Case Study}
To qualitatively assess our framework, we present two analyses: attention visualization and prediction comparison.

\noindent\textbf{Prediction Visualization.}
Figure~\ref{fig:casestudy} compares sentiment predictions with and without anchors. In the first case (\textit{``and I hated home on the range''}), the full SSU model outputs a score of \textbf{-2.21} (vs. ground truth \textbf{-2.60}), whereas the anchor-free variant yields a much weaker response (\textbf{-0.28}). With anchors, attention over text is sharply focused on sentiment-bearing words like \textit{``hated''}, and aligns closely with expressive audio and visual cues (e.g., stressed vocal segments, negative facial expressions). In the second case, the anchor enables the model to correctly associate semantically rich tokens (e.g., \textit{``Star Wars''}) with nonverbal sentiment indicators, reflecting improved generalization across modalities.

\noindent\textbf{Cross-Modal Attention Analysis.}
Figure~\ref{fig:attentionmap} shows the text-audio and text-video attention maps for the first case. Without anchors, attention is scattered and lacks clear semantic focus. With anchors, attention becomes more concentrated and sentiment tokens align with salient nonverbal regions. Red boxes highlight the top-N attention weights, indicating the strongest cross-modal connections between text and audio/video segments. These results confirm that semantic anchor sharpen inter-modal alignment and enhance interpretability.

\section{Conclusion}
We propose SSU, a unified and lightweight framework for multimodal sentiment analysis that integrates intra-modal structural modeling with cross-modal semantic alignment. SSU builds modality-specific graphs using syntax and text-guided attention, and introduces semantic anchors to align heterogeneous modalities and stabilize training. A multi-view contrastive objective further enhances semantic coherence, structural integrity, and task discriminability. Extensive experiments on CMU-MOSI and CMU-MOSEI demonstrate that SSU achieves superior accuracy and efficiency, while case studies highlight its interpretability in capturing fine-grained sentiment signals.

Overall, SSU offers a scalable and interpretable approach that effectively bridges structure and semantics in multimodal sentiment understanding.

\bibliography{aaai2026}

\end{document}